\documentclass[10pt,twocolumn,letterpaper]{article}

\usepackage{wacv}              % To produce the CAMERA-READY version

\usepackage{times}
\usepackage{epsfig}
\usepackage{graphicx}
\usepackage{amsmath}
\usepackage{amssymb}
\usepackage{booktabs}
\usepackage{multirow}

\usepackage[pagebackref,breaklinks,colorlinks]{hyperref}
\usepackage[capitalize]{cleveref}
\crefname{section}{Sec.}{Secs.}
\Crefname{section}{Section}{Sections}
\Crefname{table}{Table}{Tables}
\crefname{table}{Tab.}{Tabs.}

\def\wacvPaperID{1743} % *** Enter the WACV Paper ID here
\def\confName{WACV}
\def\confYear{2024}

\begin{document}

%%%%%%%%% TITLE
%\title{Robot Pose Nowcasting: distilling the future to improve the present}

%\title{Robot Pose Nowcasting: looking at the future to improve the present}

%\title{Robot Pose Nowcasting: forecast the future to improve the present}

%\title{Depth-based Robot Pose Nowcasting: improving pose estimation with short-term forecasting}

%\title{Robot Pose Nowcasting via Depth-based Multi-task Framework}

\title{3D Pose Nowcasting: Forecast the Future to Improve the Present}

\author{Alessandro Simoni$^*$, Roberto Vezzani\\
University of Modena and Reggio Emilia\\
% Via Pietro Vivarelli, 10, Modena, Italy, 41126\\
{\tt\small \{alessandro.simoni, roberto.vezzani\}@unimore.it}
\and
Francesco Marchetti$^*$, Lorenzo Seidenari, Alberto Del Bimbo\\
University of Florence\\
{\tt\small \{francesco.marchetti, lorenzo.seidenari,alberto.delbimbo\}@unifi.it}
\and
Federico Becattini\\
University of Siena\\
{\tt\small \{federico.becattini\}@unisi.it}
\and
Guido Borghi\\
University of Bologna\\
{\tt\small \{guido.borghi\}@unibo.it}
}
\maketitle

%%%%%%%%% ABSTRACT
\begin{abstract}
    Technologies to enable safe and effective collaboration and coexistence between humans and robots have gained significant importance in the last few years. A critical component useful for realizing this collaborative paradigm is the understanding of human and robot 3D poses using non-invasive systems. Therefore, in this paper, we propose a novel vision-based system leveraging depth data to accurately establish the 3D locations of skeleton joints. Specifically, we introduce the concept of Pose Nowcasting, denoting the capability of the proposed system to enhance its current pose estimation accuracy by jointly learning to forecast future poses. The experimental evaluation is conducted on two different datasets, providing accurate and real-time performance and confirming the validity of the proposed method on both the robotic and human scenarios.
\end{abstract}

%%%%%%%%% BODY TEXT
\section{Introduction} \label{sec:introduction} % Guido e Roberto

% \red{
% \begin{itemize}
%     %\item motivazione robot pose estimation e forecasting non chiara a una prima lettura (R1)
%     %\item togliere concetto di multi-task
%     %\item aggiungere lista contribution
%     %\item \green{figura 2 togliere xyz image e rinominare GRU in "motion encoder"}
%     \item motivare linear (aggiunto modello senza backbone in tabella 2)
%     %\item chiarire differenza tra prediction e forecasting
% \end{itemize}
% }

We are increasingly approaching an era in which humans and robots will share different spaces and moments of the day, both in social and working scenarios~\cite{peshkin2001cobot}.

Non-invasive camera monitoring combined with specific computer vision algorithms, such as Robot and Human Pose Estimators~\cite{zheng2023deep,lee2020camera}, are key and enabling technologies for safe interaction between humans and robots~\cite{colgate2008safety}. %\eg collaborative robots.
For instance, in the Industry 4.0 setting~\cite{lasi2014industry}, in which the same workplace is shared between workers and cobots~\cite{kolbeinsson2019foundation}, the ability to detect poses and avoid collisions is fundamental for safety. 
Furthermore, recent investigations~\cite{weiss2011exploring, weiss2021cobots} confirm that -- rather than the complete removal of humans~-- future generations of manufacturing will support the coexistence of humans and cobots, stressing the urgency for new investigations related to physical and social coworker coordination~\cite{dautenhahn2011new}.
Another possible application setting is represented by home automation, in which robots can autonomously perform actions but also interact with humans. 

In both cases, technologies based on non-invasive sensors that are agnostic with respect to the state of the robot's encoders, are highly desirable. 
A variety of collision detection systems, especially for the industrial environment, has been proposed but, unfortunately, they often require the use of specific sensors~\cite{hasegawa2010development}, markers~\cite{kalaitzakis2021fiducial} or access to the robot's proprietary software~\cite{geravand2013human}, which is not always possible.

\begin{figure}[t]
    \centering
    \includegraphics[width=1\columnwidth]{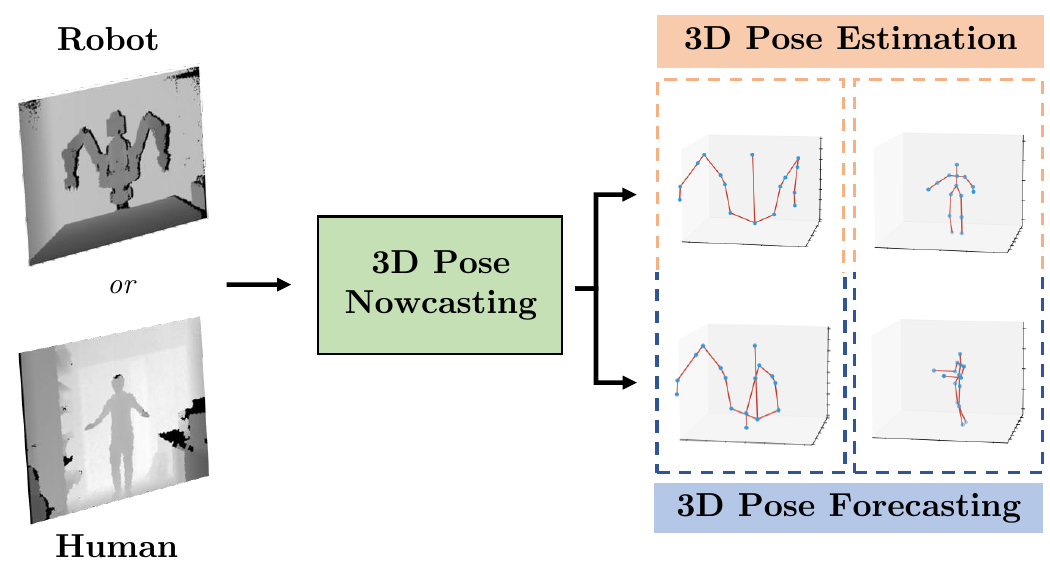}
    \caption{Estimating current and future poses through 3D Pose Nowcasting, using depth images as input data, is a fundamental technology for safe interaction between workers and collaborative machines in indoor scenarios, such as the Industry 4.0 setting.}
    \label{fig:initial}
\end{figure}

Therefore, in this paper, we propose a vision-based system able to accurately estimate the 3D poses by learning to forecast the near future as an auxiliary task.
In particular, we show how the knowledge about the future at training time improves the model's performance in the present. 

Given the similarities with the weather forecasting~\cite{browning1989nowcasting}, we refer to this novel paradigm as \textbf{3D Pose Nowcasting}, characterized by the following elements: 
i) the forecasting regards a brief time span (around a few seconds);
ii) we are not required to access specific physical models or additional sensors other than the input data (in our case, depth images);
iii) forecasting, in addition to enhancing present estimation, is important to raise alarms about imminent and unexpected events (\eg collisions, hazards).

The proposed method for 3D Pose Nowcasting, outlined in Figure~\ref{fig:initial}, is based on addressing the task from two different research fields, \ie 3D Pose Estimation (PE) and 3D Pose Forecasting (PF), jointly learned during training.
In particular, the model is trained end-to-end to estimate the 3D pose at the current timestep and the 3D poses at the next future timesteps.
%In the paper, the term ``prediction" is used to generically refer to either estimates or forecasts in the respective tasks.

Our approach is based on depth data enabling the development of a vision-based system robust to varying or absent environmental light sources~\cite{sarbolandi2015kinect}, usually common in indoor scenarios such as workplaces. Besides, depth acquisition devices nowadays are inexpensive, yet accurate~\cite{zanuttigh2016time}. 
Moreover, in the Sim2Real~\cite{hofer2021sim2real} setting, the use of depth reduces the domain gap between synthetic and real scenarios~\cite{simoni2022semi}, thus enabling the usage of large-scale datasets without the time-consuming collecting and labeling procedures required with real data.

From an architectural point of view, PE and PF are tackled through two double-branch CNNs, each specialized in estimating and forecasting joints in 3D world coordinates. 
The first branch is composed of a backbone originally developed for Human Pose Estimation~\cite{andriluka20142d} (HPE), while the second one is obtained by exploiting a motion encoder based on a recurrent neural network, that processes a sequence of past joint locations.
The 3D world-coordinate locations of each joint are given in output in real-time, leveraging the recent Semi-Perspective Decoupled Heatmaps (SPDH)~\cite{simoni2022semi} as an intermediate representation of poses. 
To train the model, a double loss is used to optimize both the current pose and the future poses. This is justified by the fact that we want the forecasting loss to influence and improve the estimate at the current timestep.
% Predictions are buffered and fed back to the temporal branch of the model to forecast future poses.
%
%The proposed system is trained and tested on a dataset, namely SimBa~\cite{simoni2022semi}, specifically created for the estimation of robotic joints, starting from a synthetic depth domain. In addition, also due to a lack of literature datasets, we seize the opportunity to test our system on the prediction of human joints, through the well-known ITOP~\cite{haque2016towards} dataset. 

% The proposed system has been trained and tested on the SimBa dataset~\cite{simoni2022semi}, specifically created for the estimation of robotic joints from depth images.
% Moreover, through SimBa, we seize the opportunity to test our system in the Sim2Real~\cite{hofer2021sim2real} scenario, in which training data consists of synthetic sequences while testing data are sequences collected through a real depth device. 
% On one side, this cross-domain scenario has proven to be challenging, on the other, it avoids the need for costly data collection and labeling procedures.
% The proposed system has been trained and tested on the SimBa dataset~\cite{simoni2022semi}, specifically created for the estimation of robotic joints from depth images.
% In addition, we demonstrate the flexibility of our approach by testing the system on the ITOP~\cite{haque2016towards} dataset, which has characteristics similar to the context of our interest, albeit applied to human poses.

Summarizing, the main contributions of our paper are:
\begin{itemize}
    \item We introduce the novel paradigm of 3D Pose Nowcasting, a combination of 3D Pose Estimation and 3D Pose Forecasting in a joint optimization framework. By learning to predict the future, our model improves its pose estimation accuracy in the present.
    \item We demonstrate the robustness of our approach in the Sim2Real scenario, enabling effective exploitation of synthetic data at training time, and also domain transfer capabilities from synthetic to real.
    \item We obtain state-of-the-art performance in estimating the current robot's pose, also providing reliable future predictions. In addition, we show that 3D Pose Nowcasting can be easily exploited for estimating human body joints.
\end{itemize}

% Experimental results confirm that: 
% i) our system achieves state-of-the-art performance in predicting the current robot's pose; 
% ii) it also predicts near-future joint locations with reliable precision, in a time range of $2$ seconds;  
% iii) interestingly, the forecasting of future joint positions also helps to improve the performance of the RPE task.

\begin{figure*}[th!]
    \centering
    \includegraphics[width=1\textwidth]{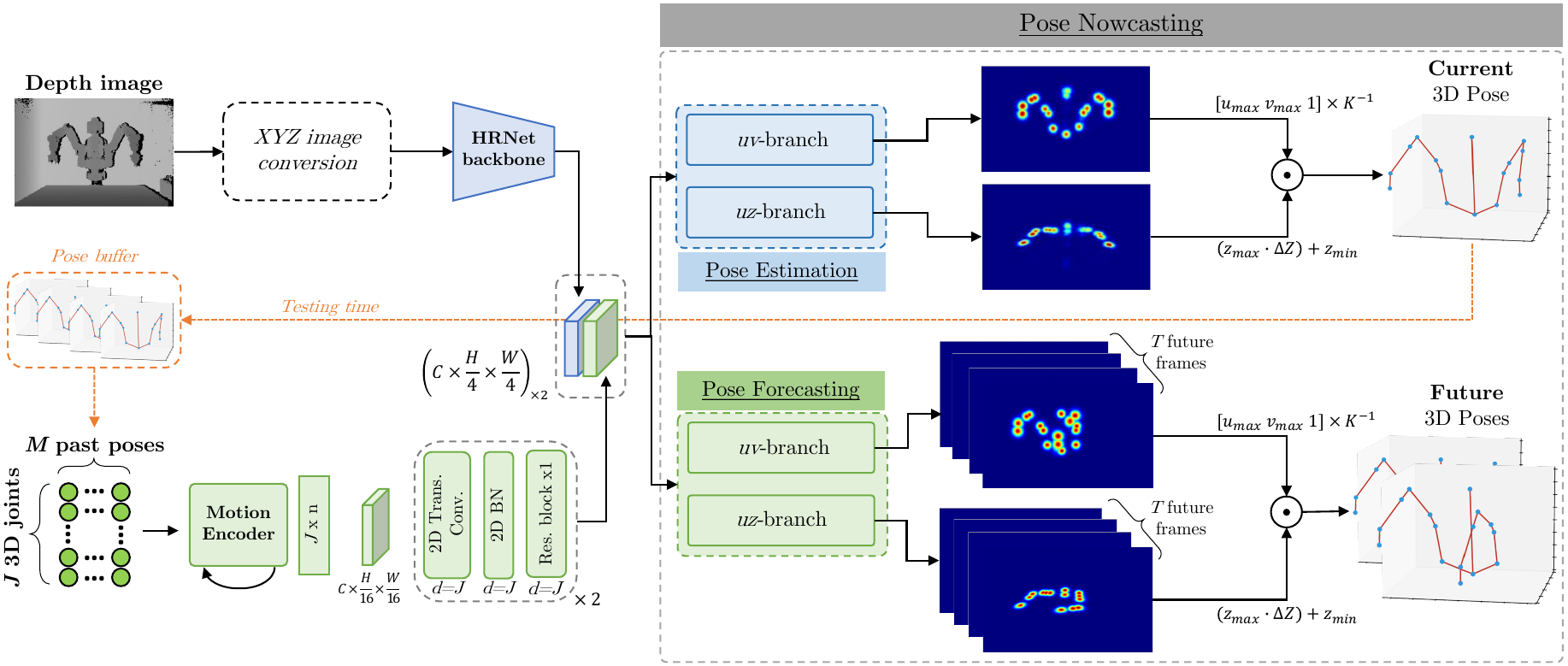}
    \caption{Overview of the proposed 3D Pose Nowcasting framework. First, features related to the depth map and the past poses are extracted. These features are then concatenated and fed to two different branches, \ie the Pose Estimation and Pose Forecasting ones. Finally, the framework outputs the current and the near-future 3D poses. For the sake of visualization, heatmaps are stacked channel-wise.}
    \label{fig:general}
\end{figure*}

% Summarizing, the contributions of the paper are:
% \begin{itemize}

%     \item We propose the first multi-task nowcasting framework that estimates and forecasts the 3D robot's joint locations with state-of-the-art accuracy. We observe that it is an enabling non-invasive technology for safe interaction between humans and collaborative robots.

%     \item The presented system is based only on depth images, a type of data not fully investigated in the task of robot pose estimation and forecasting. This sensor choice makes our vision-based system more robust to light changes in source and intensity, relying on accurate and cheap devices.
    
% \end{itemize}

\section{Related Work}

\textbf{Robot Pose Estimation from Depth.}
Only a limited amount of research addresses the task of pose estimation from depth data. 
% Taking inspiration from the HPE field~\cite{shotton2012efficient}, 
Bohg \etal.~\cite{bohg2014robot} proposed to use a random forest classifier to classify and then group depth maps pixels, obtaining skeleton joints. A similar approach is reported in~\cite{widmaier2016robot}, in which joint angles are directly regressed without any segmentation prior.
However, these methods are unable to infer real-world 3D poses, limiting their estimates to joint angles.
The large majority of literature works for robot pose estimation are developed for the RGB domain.
% For the sake of completeness, we also briefly investigate research in the RGB domain.
In general, there are two main approaches: 
% there are two main approaches for robot estimation from intensity data: 
hand-eye calibration-based and rendering-based. In the former, methods are based on fiducial markers (\eg ArUco~\cite{garrido2014automatic})
% , ARTag~\cite{fiala2005artag} and AprilTag~\cite{olson2011apriltag}), usually 
placed on the robot's end effector, tracked through multiple cameras. Then, a 3D-2D correspondence problem is solved by relying on forward kinematics or the PnP~\cite{lepetit2009epnp} approach. Unfortunately, these methods are invasive since they require the physical application of markers on the robot, which is not always feasible or practicable.
% Other hand-eye calibration-based approaches are based on the solution of the Perspective-n-Point (PnP)~\cite{lepetit2009epnp} problem, with which is possible to retrieve 3D coordinates starting from the 2D pose of the robot and the robot's model.
Differently, rendering-based methods~\cite{labbe2021single,noguchi2022watch} use the render\&compare paradigm, where an optimization algorithm iteratively refines the pose projected to the image with respect to the camera. 
% One of the first approaches with this technique is described in~\cite{labbe2021single}: at test time, the approach can also work with unknown joint angles, but with a significant drop in performance. In~\cite{noguchi2022watch}, a self-supervised method is presented, based on an approximation of the robot model and its implicit refinements. 

% Differently from the aforementioned approaches, the proposed method is not based on the state of the robot's encoders and on multiple view acquisition, simplifying the deployment to real-world scenarios.

\textbf{Human Pose Estimation from Depth.}
Shotton et al.~\cite{shotton2012efficient} introduced a pioneering approach based on a random forest classifier to classify pixels 
% in input depth maps, 
enabling the segmentation of the human body. The 3D joint candidates are then identified through a weighted density estimator. 
Using similar features, in~\cite{yub2015random} the authors proposed to use a regression tree to predict the probability distribution of the direction of a specific joint. 
% This approach is able to run up to 1000 fps.
Entering the deep learning-based field, some works introduce the use of NNs in combination with a single depth frame. In~\cite{wang2018convolutional}, a specific memory module referred to as Convolutional Memory Block is introduced, merging the power of CNNs and a memory mechanism used to handle depth data.
More recently, \cite{garau2021deca} introduced a capsule autoencoder network based on fast Variational Bayes capsule routing, focusing on improving viewpoint generalization both on intensity and depth data.
Other works are based on point clouds sampled from depth data. In particular, the method described in~\cite{zhang2020weakly} is based on a point clouds proposal module followed by a 3D pose regression module. Similarly, the same authors in~\cite{zhang2021sequential} introduced a sequential pose estimation module based on a window of different frames, improving the general performance at the cost of increasing computational complexity.
Finally, some literature works have been developed originally for the hand pose estimation task~\cite{moon2018v2v,xiong2019a2j,guo2017towards} and then adapted to tackle also the human pose estimation task.

\textbf{Pose Forecasting.}
% Despite the growing interest in this kind of task, only a few works address forecasting poses of industrial robots. In fact, most works focus on analyzing humans rather than robots, even when the two are supposed to cooperate with each other~\cite{dallel2020inhard}. 
Recently, Sampieri \etal~\cite{sampieri2022pose} proposed a graph convolutional neural network to jointly model robot arms and human operators from RGB images. Their goal is to anticipate human-robot collisions. In this work, we follow this research direction and we leverage a trajectory forecasting architecture to improve the current 3D robot pose estimate while also providing information about the future locations of robots and humans.
From a general point of view, a large crop of literature has addressed motion forecasting tasks, especially in automotive~\cite{lee2017desire, marchetti2020multiple, ivanovic2019trajectron, luc2018predicting} and human behavior understanding~\cite{pavllo2018quaternet, toyer2017human, chiu2019action, chao2017forecasting, diller2022forecasting}.
The task can be framed as an encoder-decoder problem, where past motion is projected into a latent state and then decoded into a plausible future~\cite{lee2017desire, alahi2016social}.
Interestingly, most approaches formulate the forecasting task as a multimodal prediction task, due to the intrinsic uncertainty of the problem~\cite{vondrick2016anticipating, lee2017desire, salzmann2022motron, guimard2022deep}. 
% Most works have traditionally focused on forecasting trajectories for vehicles \cite{lee2017desire, marchetti2020multiple, srikanth2019infer, chang2019argoverse} and pedestrians \cite{marchetti2022smemo,pellegrini2009you,salzmann2020trajectron++}.
% Model-wise, an array of architectures have proven effective for this task such as LSTMs \cite{alahi2016social}, GANs \cite{gupta2018social, amirian2019social}, graph neural networks \cite{Kosaraju2019BIGAT,mohamed2020social,shi2021sgcn}, memory augmented networks \cite{marchetti2020mantra, marchetti2022smemo, xu2022remember}, variational autoencoders \cite{lee2017desire} and transformers \cite{giuliari2021transformer}.
% These works approximate the overall motion of the agents as 2D points, observed in a bird-eye view map or forecast spatial positions as bounding boxes from an egocentric point of view.
More recently, several works have addressed the task of forecasting human poses. Compared to the automotive setting, this is a much more complex scenario, since body joints can move erratically and the position of the whole skeleton must be predicted at every timestep. Here, graph-based representations play an important role, since body joints can be naturally represented as connected nodes \cite{plizzari2021spatial, li2021gpfs, adeli2021tripod, sofianos2021space}.
Unlike these methods, Mangalam \etal \cite{mangalam2020disentangling} fused 3D skeletons, camera ego-motion and monocular depth estimates to forecast body poses. In a similar way, we propose a depth-based approach for pose estimation and forecasting. Differently from \cite{mangalam2020disentangling}, we focus on robot poses and, instead of observing a full sequence of depth and joints, we blend the current depth with an encoding of autoregressively generated past joints.

\textbf{Depth-based datasets for Pose Estimation and Forecasting.} 
We observe a substantial lack of datasets that can be used for robot pose estimation and forecasting starting from depth data. Recently, four different datasets have been introduced in the literature, but totally based on RGB data. Released in 2019, the CRAVES~\cite{zuo2019craves} dataset consists of synthetic and real acquisitions of a single type of robotic arm, for a total of about $5$k frames. DREAM~\cite{lee2020camera} and WIM~\cite{noguchi2022watch}, introduced in 2020 and 2022, contain $350$k and $140$k intensity frames, respectively, depicting different types of robots. One of the most recent datasets is referred to as CHICO~\cite{sampieri2022pose}. Expressively introduced for collision detection in human-robot interaction, it collects more than $1$ million frames acquired with multiple RGB cameras~\footnote{This dataset presents corrupted 3D joint annotations on images not yet fixed by the authors, making it impossible for us to adopt it.}. Therefore, the only dataset exploitable to test our method is the recent SimBa~\cite{simoni2022semi}, consisting of more than $370$k frames depicting the Rethink Baxter robot performing pick-and-place operations in random locations. This dataset has been acquired in the Sim2Real~\cite{hofer2021sim2real} scenario, \ie the training and testing frames belong to two different domains: synthetic (generated through ROS and Gazebo~\cite{koenig2004design} simulator) and real (acquired through the time-of-flight Microsoft Kinect v2 depth device). SimBa is suitable for our task due to the presence of video sequences, collected at $30$ fps.

With regard to the estimation of human poses, we adopt the ITOP dataset~\cite{haque2016towards}, which has been used as a benchmark by several prior works \cite{zhang2020weakly, garau2021deca, zhang2021sequential, wang2018convolutional, garau2023capsules}. Also in this case, we observe a substantial lack of depth-based datasets in the literature, suitable for our method, for different motivations.
Human3.6M~\cite{ionescu2013human3} dataset contains very low-quality depth images, acquired through the MESA Imaging SR4000 device. 
The NTU dataset~\cite{trivedi2021ntu}, originally developed for the human action recognition task, contains good quality depth data, but unfortunately, the human pose annotations are automatically provided through the method described in~\cite{shotton2012efficient}, reducing their accuracy. 
The mRI dataset~\cite{an2022mri} appears to be an interesting dataset but depth data have yet to be released, at the time of writing.

\section{Proposed Method} % Lorenzo e Federico
% c'era molta confusione con i nomi. Ho cercato di uniformare: input branch, output branch (ognuno dei quali con due sub-branch), nowcasting block
% An overview of the proposed framework is depicted in Figure \ref{fig:general} and it is mainly organized into two input branches, two output branches, each consisting of two uv/uz sub-branches, that compose the Pose Nowcasting block. 
% In this organization, the first couple of input and output branches (blue) is developed for the pose estimation task, while the second couple (green) is focused on pose forecasting.

An overview of the proposed framework is depicted in Figure \ref{fig:general}. It is organized in an encoder-decoder fashion that is split into two input branches and two output branches. The encoder extracts visual and temporal embeddings, while the decoder consists of the \textit{Pose Nowcasting} block, which is made of two SPDH~\cite{simoni2022semi} branches dedicated to pose estimation and pose forecasting.
% Visually, the first couple of input and output branches (blue) is developed for the pose estimation task, while the second couple (green) is focused on pose forecasting.

From a formal point of view, the encoder can be viewed as a single frame 2D depth input branch $\Pi(\cdot)$ and a temporal 3D joint recurrent input branch $\Gamma(\cdot)$. For a depth image $D$ and a sequence of $t=1,...,M$ poses $P_j^t = [X_j^t, Y_j^t, Z_j^t]$ with $j=1,...,J$ 3D joints, two same-size feature maps $\Pi(D)$ and $\Gamma(\mathbf{P})$ are computed and concatenated. 
The output branches of the nowcasting decoder then independently generate current and future pose predictions.

% \todo{Fede: ho cambiato il numero di joints K in J, visto che K è la matrice di proiezione. Di conseguenza ho cambiato J in $\Gamma$ così da uniformare la notazione per i due branch, entrambi con lettere greche. Occhi aperti per vecchie notazioni che potrebbero essermi sfuggite nel testo. Fixare l'immagine di conseguenza.}

\subsection{Depth and Past Pose Input Processing}
As mentioned, the first input branch is responsible for extracting the features related to the current pose. In this case, the input is represented by a depth image that is converted into an XYZ image, formally defined as follows:
\begin{equation}
    I_{XYZ} = \pi(D \cdot K^{-1})
\end{equation}
where $\pi$ is the projection in the 3D space, $D$ is the matrix of distances used to create the depth image and $K$ is the projection matrix. This kind of depth representation has been proved to have better generalization capabilities across different domains with respect to common depth images \cite{simoni2022semi}.
Being aware of the recent and significant advances in HPE~\cite{dang2019deep}, we exploit the well-known HRNet-32 architecture~\cite{sun2019deep}, specifically the randomly initialized first four stages without the last convolution, as the backbone to extract pose-related features. These features are then concatenated with the ones extracted through the other branch, described as follows.

%The second part of the system receives as input a set of the 3D joint positions that belong to the last $T$ frames. These joints are then fed into a Gated Recurrent Unit (GRU), followed by a layer with  

% Similarly to Human Pose Estimation frameworks our framework output joint predictions as heatmaps, in our case we predict  decoupled uz/uv maps. 
The second input branch incorporates temporal information obtained from previously estimated 3D joint positions: this information becomes available as soon as a buffer of poses of length $M$ is filled by storing the outputs of the pose estimation block.
This branch uses a motion encoder, implemented as a GRU\footnote{Potentially any kind of recurrent architecture such as LSTMs or Transformers could be used. Since our focus is on Nowcasting, we adopt GRUs as commonly done in the trajectory forecasting literature, leaving the investigation of different architectures to future research.}, to process higher dimensional embeddings of each pose $P_j^t$. Its output is organized into a $C\times \frac{H}{16}\times \frac{W}{16}$ shaped feature map, which is then processed with two layers of residual transposed convolutions with BatchNorm. 
This architecture is both responsible for processing temporal information stored in previously estimated joints and for adapting the 3D representation to a 2D map that can be fused with the feature map extracted by $\Pi(\cdot)$ from depth images.

\begin{figure}[t]
    \centering
    \includegraphics[width=1\columnwidth]{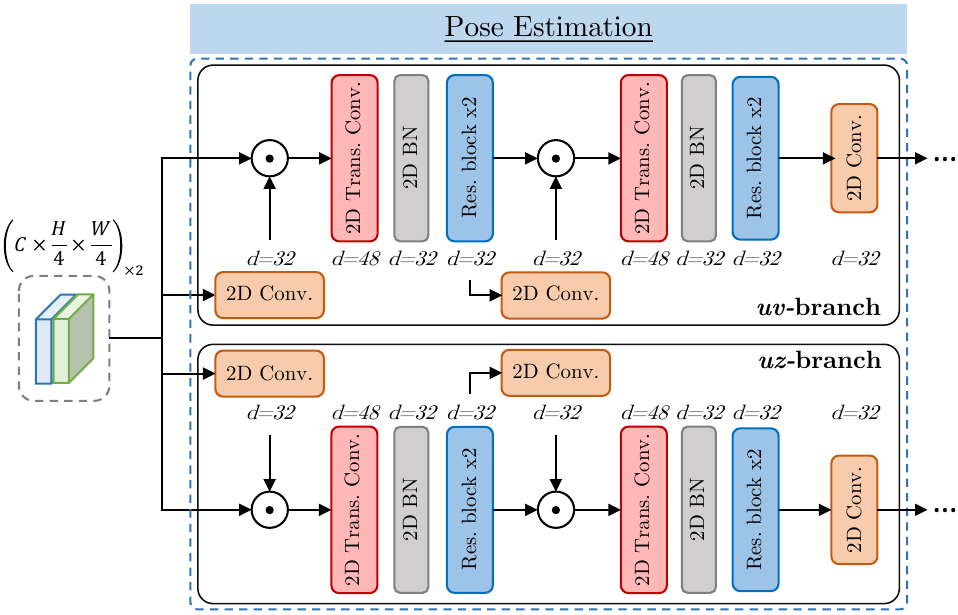}
    \caption{Architecture of the Pose Estimation branch. The input is represented by the concatenation of features extracted from depth maps and past joints. Each \textit{uv}/\textit{uz} sub-branch generates the heatmap-based SPDH~\cite{simoni2022semi} representation of 3D joint locations.}
    \label{fig:spdh}
\end{figure}

\begin{table*}[t]
    \begin{center}
    \resizebox{0.84\linewidth}{!}{
    \begin{tabular}{l r c ccccc c c}
    \toprule
        & & & \multicolumn{5}{c}{\textbf{mAP} (\%) $\uparrow$} & & \\
        \cmidrule{4-8}
        \textbf{Input} & \textbf{Model} & & $\mathbf{2}$cm & $\mathbf{4}$cm & $\mathbf{6}$cm & $\mathbf{8}$cm & $\mathbf{10}$cm & & \textbf{ADD} (cm) $\downarrow$ \\
        \midrule
        Depth & ResNet-18~\cite{he2016deep} & & $0.57$ & $9.40$ & $19.99$ & $27.06$ & $44.44$ && $12.20 {\scriptstyle \, \pm 4.12}$ \\
        2D joints & Martinez \etal~\cite{martinez2017simple}~$^{*}$ & & $13.70$ & $26.96$ & $37.98$ & $48.40$ & $58.33$ && $10.03 {\scriptstyle \, \pm 3.53}$ \\
        Depth & Pavlakos \etal~\cite{pavlakos2017coarse} & & $3.35$ & $18.15$ & $42.24$ & $61.60$ & $86.15$ && $7.11 {\scriptstyle \, \pm 0.65}$ \\
        Depth & Simoni \etal~\cite{simoni2022semi} & & $6.33$ & $53.75$ & $79.75$ & $93.90$ & $98.12$ & & $4.41 {\scriptstyle \, \pm 1.09}$ \\
        \midrule
        Depth & \textit{Ours w/o forecasting} & & $16.25$ & $57.51$ & $89.81$ & $\mathbf{99.26}$ & $\mathbf{99.81}$ & & $3.77 {\scriptstyle \, \pm 0.98}$ \\
        Depth + $M$ past poses & \textit{Ours} & & $\mathbf{30.68}$ & $\mathbf{66.90}$ & $\mathbf{92.69}$ & $98.02$ & $98.38$ & & $\mathbf{3.52 {\scriptstyle \, \pm 1.30}}$ \\
        \bottomrule
    \end{tabular}
    }
    \end{center}
    \caption{Robot pose estimation results on SimBa. The proposed framework is tested by taking as input a single depth image (``\textit{Ours w/o forecasting}'') or a depth image with the previously predicted 3D joints (``\textit{Ours}''). Method marked with $^*$ uses a relative joint representation.}
    \label{tab:simba-3d-results}
\end{table*}

\subsection{Pose Estimation and Forecasting Branches}
Our framework is completed by the nowcasting block with two output branches jointly solving pose estimation and forecasting.
Both branches exploit the same SPDH~\cite{simoni2022semi} representation, in which the 3D space is decomposed into two bi-dimensional spaces where skeleton joint locations are expressed through heatmaps. In particular, the $uv$ space corresponds to the camera image plane (the front view of the acquired scene), while the $uz$ space contains the quantized values of the depth dimension, \ie a sort of birds-eye view of the scene with discretized information about the distance of the joints. 

In the \textit{pose estimation branch}, the SPDH representation is obtained through the architecture detailed in Figure~\ref{fig:spdh}, consisting of two residual transposed convolution layers followed by a BatchNorm and ReLU activation function. The estimated pose is represented by a set of $J \times 2$ 
%$uv/uz$ 
heatmaps, one pair for each joint in the $uv$ and $uz$ spaces.

In the \textit{pose forecasting branch}, we adopt a lighter architecture to deal with the multiple SPDH representations that aim to model the near-future joint locations.
In particular, we use two 2D convolutional layers, with a size of $32$, interspersed with a BatchNorm and ReLU activation function.
The forecasted poses are represented as $T \times (J \times 2)$ future 
%$uv$/$uz$ 
heatmaps, where $T$ is the forecasting horizon.

For both output branches, final predictions are obtained as follows: we compute the argmax of the $uv$ heatmaps and we multiply the resulting values ($u_{max}, v_{max}$) with the inverse of the camera intrinsics $K^{-1}$ to obtain the final 3D coordinates.
Differently, with $uz$ heatmaps, we transform the result of the argmax operation into a continuous value in the metric space multiplying it with the quantization step ($\Delta Z$) computed in the defined depth range ($z_{min}, z_{max}$).
% We convert heatmaps into real-world coordinates using camera intrinsics. Moreover, we solve the more challenging task of pose forecasting up to 2s in the future with a multi-frame output head.

\subsection{Losses}
To train the model, we directly optimize the $uv/uz$ heatmaps, before they are converted into 3D coordinates. The system is trained end-to-end optimizing the Mean Squared Error (MSE) loss function $\mathcal{L}$ between generated and ground truth heatmaps:
% \begin{equation}\label{eq:loss}
%     \mathcal{L}= \frac{1}{|\mathcal{J}|}\sum_{j \in \mathcal{J}} \bigg( || H_j^t - \widehat{H}_j^t ||_2 + \frac{1}{T}\sum_{k=t}^{t+{T}} ||H_j^k - \widehat{H}_j^k ||_2\bigg)
% \end{equation}

\begin{equation}\label{eq:loss_pe}
    \mathcal{L}_{PE} = \frac{1}{|\mathcal{J}|}\sum_{j \in \mathcal{J}} || H_j^t - \widehat{H}_j^t ||_2
\end{equation}
\begin{equation}\label{eq:loss_pf}
    \mathcal{L}_{PF} = \frac{1}{|\mathcal{J}|}\sum_{j \in \mathcal{J}} \frac{1}{T}\sum_{k=1}^{t+{T}} ||H_j^{t+k} - \widehat{H}_j^{t+k} ||_2
\end{equation}

\begin{equation}\label{eq:loss}
    \mathcal{L} = \mathcal{L}_{PE} + \mathcal{L}_{PF}
\end{equation}

where $\mathcal{L}_{PE}$ is the pose estimation loss between the estimated pose $\widehat{H}_j^t$ and the ground truth $H_j^t$ at the current timestep $t$; $\mathcal{L}_{PF}$ is the auxiliary pose forecasting loss between the sequence of $k=1,...,T$ generated future poses $H_j^{t+k}$ and their corresponding ground truths $\widehat{H}_j^{t+k}$; and $\mathcal{J}$ is the set of skeleton joints in both the \textit{uv} and \textit{uz} views.
%and ${T}$ is the amount of timestamps predicted in the future.
Note that $\widehat{H}_j^t$ is generated by the pose estimation branch whether $\widehat{H}_j^{t+k}$ are generated by the pose forecasting branch.

%where $H_j^t$ and $\widehat{H}_j^t$ are the ground truth and predicted heatmaps for joint $j$ at time $t$ respectively, and $\mathcal{J}$ is the set of robot joints in both the \textit{uv} and \textit{uz} views and ${T}$ is the amount of timestamps predicted in the future.}

\begin{table*}[t]
    \begin{center}
    \resizebox{0.84\linewidth}{!}{
    \begin{tabular}{l c l c ccccc c c}
    \toprule
        & & & & \multicolumn{5}{c}{\textbf{mAP} (\%) $\uparrow$} & & \\
        \cmidrule{5-9}
        \textbf{Input} & \textbf{Model} & \textbf{Horizon} & & $\mathbf{2}$cm & $\mathbf{4}$cm & $\mathbf{6}$cm & $\mathbf{8}$cm & $\mathbf{10}$cm & & \textbf{ADD} (cm) $\downarrow$ \\
        \midrule
        $M$ past poses & \textit{Linear} & $t$ & & $0.31$ & $6.02$ & $15.81$ & $25.78$ & $41.23$ & & $ 16.89{\scriptstyle \, \pm  5.73}$ \\
        $M$ past poses & \textit{Linear} & $t+0.5s$ & & $0.42$ & $5.54$ & $15.34$ & $25.40$ & $40.58$ & & $ 17.54 {\scriptstyle \, \pm 6.20}$ \\
        $M$ past poses & \textit{Linear} & $t+1s$ & & $0.29$ & $4.78$ & $14.76$ & $23.44$ & $38.08$ & & $19.25 {\scriptstyle \, \pm 6.20}$ \\
        $M$ past poses & \textit{Linear} & $t+1.5s$ & & $0.32$ & $4.34$ & $14.11$ & $22.76$ & $36.72$ & & $19.75 {\scriptstyle \, \pm 6.17}$ \\
        $M$ past poses & \textit{Linear} & $t+2s$ & & $0.37$ & $3.98$ & $13.72$ & $21.84$ & $35.96$ & & $ 20.04{\scriptstyle \, \pm 6.10}$ \\
        \midrule
        $M$ past poses & \textit{Ours} & $t$ & & $5.33$ & $22.77$ & $37.42$ & $57.96$ & $78.05$ & & $8.38 {\scriptstyle \, \pm 3.88}$ \\
        $M$ past poses & \textit{Ours} & $t+0.5s$ & & $4.77$ & $20.74$ & $37.31$ & $55.63$ & $76.53$ & & $8.61 {\scriptstyle \, \pm 4.07}$ \\
        $M$ past poses & \textit{Ours} & $t+1s$ & & $4.41$ & $19.65$ & $35.58$ & $53.16$ & $73.58$ & & $9.09{\scriptstyle \, \pm 4.04}$ \\
        $M$ past poses & \textit{Ours} & $t+1.5s$ & & $4.12$ & $19.34$ & $33.40$ & $51.65$ & $72.08$ & & $9.73{\scriptstyle \, \pm 4.23}$ \\
        $M$ past poses & \textit{Ours} & $t+2s$ & & $4.02$ & $18.81$ & $32.56$ & $50.32$ & $70.21$ & & $10.41{\scriptstyle \, \pm 4.59}$ \\
%        \midrule
%        Depth + $M$ past poses & \textit{Ours} & $t$ & & $30.68$ & $66.90$ & $92.69$ & $98.02$ & $98.38$ & & $\mathbf{3.52 {\scriptstyle \, \pm 1.30}}$ \\
%        Depth + $M$ past poses & \textit{Ours} & $t+0.5s$ & & $31.32$ & $66.04$ & $91.71$ & $97.66$ & $98.33$ & & $\mathbf{3.57 {\scriptstyle \, \pm 1.33}}$ \\
%        Depth + $M$ past poses & \textit{Ours} & $t+1s$ & & $28.89$ & $59.67$ & $84.39$ & $91.04$ & $92.65$ & & $\mathbf{4.50 {\scriptstyle \, \pm 2.25}}$ \\
%        Depth + $M$ past poses & \textit{Ours} & $t+1.5s$ & & $26.41$ & $55.99$ & $78.14$ & $85.93$ & $87.93$ & & $\mathbf{5.71 {\scriptstyle \, \pm 3.48}}$ \\
%        Depth + $M$ past poses & \textit{Ours} & $t+2s$ & & $25.04$ & $53.43$ & $73.52$ & $81.27$ & $83.39$ & & $\mathbf{6.85 {\scriptstyle \, \pm 4.38}}$ \\
%%%%all bold ours
\midrule
Depth + $M$ past poses & \textit{Ours} & $t$ & & \textbf{30.68} & \textbf{66.90} & \textbf{92.69} & \textbf{98.02} & \textbf{98.38} & & $\mathbf{3.52 {\scriptstyle \, \pm 1.30}}$ \\
Depth + $M$ past poses & \textit{Ours} & $t+0.5s$ & & \textbf{31.32} & \textbf{66.04} & \textbf{91.71} & \textbf{97.66} & \textbf{98.33} & & $\mathbf{3.57 {\scriptstyle \, \pm 1.33}}$ \\
Depth + $M$ past poses & \textit{Ours} & $t+1s$ & & \textbf{28.89} & \textbf{59.67} & \textbf{84.39} & \textbf{91.04} & \textbf{92.65} & & $\mathbf{4.50 {\scriptstyle \, \pm 2.25}}$ \\
Depth + $M$ past poses & \textit{Ours} & $t+1.5s$ & & \textbf{26.41} & \textbf{55.99} & \textbf{78.14} & \textbf{85.93} & \textbf{87.93} & & $\mathbf{5.71 {\scriptstyle \, \pm 3.48}}$ \\
Depth + $M$ past poses & \textit{Ours} & $t+2s$ & & \textbf{25.04} & \textbf{53.43} & \textbf{73.52} & \textbf{81.27} & \textbf{83.39} & & $\mathbf{6.85 {\scriptstyle \, \pm 4.38}}$ \\

%%%%end all bold ours
        
        % \midrule
        % Pred. 3D joints & \textit{Copy last} & $t$ & & $2.72$ & $21.95$ & $34.54$ & $47.86$ & $71.34$ & & $10.93 {\scriptstyle \, \pm 5.36}$ \\
        % Pred. 3D joints & \textit{Copy last} & $t+0.5$ & & $$ & $$ & $$ & $$ & $$ & & ${\scriptstyle \, \pm }$ \\
        % Pred. 3D joints & \textit{Copy last} & $t+1$ & & $$ & $$ & $$ & $$ & $$ & & $ {\scriptstyle \, \pm }$ \\
        % Pred. 3D joints & \textit{Copy last} & $t+1.5$ & & $$ & $$ & $$ & $$ & $$ & & $ {\scriptstyle \, \pm }$ \\
        % Pred. 3D joints & \textit{Copy last} & $t+2$ & & $$ & $$ & $$ & $$ & $$ & & $ {\scriptstyle \, \pm }$ \\
        \bottomrule
    \end{tabular}
    }
    \end{center}
    \caption{Results on both robot pose estimation and forecasting on SimBa. The proposed method is compared to a linear model and our model without the depth-based input branch, while tested in an autoregressive manner.}
    \label{tab:simba-all-results}
\end{table*}

\section{Experimental Validation}

\subsection{Datasets}
\noindent \textbf{SimBa}~\cite{simoni2022semi} is a recent dataset specifically acquired for the robot pose estimation task from depth data. It presents unique features such as the presence of synthetic and real depth data, acquired with Gazebo and the Microsoft Kinect v2 sensor.
Both domains consist of several sequences of random pick-and-place operations, acquired through randomly placed cameras (left, right and center). The acquired depth data leverages the Time-of-Flight technology and has a spatial resolution of $510 \times 424$.
This dataset has challenges due to different domains for training and testing (Sim2Real scenario) and different positions of the acquisition devices.

\noindent \textbf{ITOP}~\cite{haque2016towards} consists of $20$ subjects performing $15$ different complex actions, for a total of $50$k frames ($40$k training and $10$k testing, as reported in the original paper). Two Structured Light (SL) depth sensors (Asus Xtion Pro) are used to acquire data, one placed in front of the subject, and one placed on the top: in this paper, we focus on the side view, in which human joints are not fully occluded by the head and shoulders of the subject. 
Annotations consist of 2D and 3D joint coordinates, manually refined to lie inside the body to address human pose estimation from depth data. Unfortunately, not all annotations are valid, thus limiting the length of temporally consistent sequences.
The challenges of this dataset are related to the limited quality of depth data, in terms of spatial resolution ($320 \times 240$), depth accuracy (SL technology~\cite{sarbolandi2015kinect}), and action complexity, with several occlusions produced during movements.

The proposed system has been trained and tested on the SimBa dataset~\cite{simoni2022semi}, specifically created for the estimation of robotic joints from depth images.
In addition, we demonstrate the generalization capabilities of our approach by testing the system on the ITOP~\cite{haque2016towards} dataset, which has characteristics similar to the context of our interest, albeit applied to human poses.

\subsection{Metrics}
For the 3D pose estimation and forecasting tasks, we exploit standard literature metrics, \ie Average Distance metric (ADD) and mean Average Precision (mAP).
The first, that is the L$_2$ distance expressed in centimeters of all 3D robot joints to their ground truth positions, conveys the error related to the translation and rotation in the 3D world (the lower the better).
The second metric is defined as:
\begin{equation} \label{eq:itop_accuracy}
    \text{mAP} = \frac{1}{|N|} \, \sum_{j \in N} \, \big( \lVert \mathbf{v}_j-\widehat{\mathbf{v}}_j \rVert_2 < \delta \big)
\end{equation}
where $N$ is the number of skeleton joints, $\mathbf{v}_j$ is the predicted joint and $\widehat{\mathbf{v}}_j $ is the ground truth.
This metric is intended as the accuracy of the ADD using different thresholds ($\delta = \{2, 4, 6, 8, 10\}$ centimeters in our experiments and it improves the interpretability of results. 

\subsection{Training}
The proposed model is trained for $30$ epochs by exploiting the MSE loss for the heatmaps produced by both the branches for the current and future poses. We use the Adam optimizer, with an initial learning rate of $10^{-3}$, a decay factor of $10^{-1}$ at $50\%$ and $75\%$ of the training procedure and a batch size of $16$.
In all experiments, we use the original dataset splits to train and test the model.

During the training on both datasets, we apply data augmentation on the point clouds computed from the input depth maps.
Specifically, 3D points are randomly translated with a maximum range of $[-20\text{cm}, +20\text{cm}]$ and $[-30\text{cm}, +30\text{cm}]$ for XY and Z axes, respectively. Moreover, the points are rotated with a range of $[-5^{\circ}, +5^{\circ}]$ for the XZ axes.
In terms of visual appearance, we introduce a pepper noise on about $15\%$ of the pixels and a random dropout, consisting in setting with the value $0$ several small portions of the input image: in this manner, we simulate the presence of depth noise, usually found in real-world depth sensors, and the presence of non-reflecting surfaces (on which the depth value is not valid) in the acquired scene.

\begin{table*}[t]
    \begin{center}
    \resizebox{0.95\linewidth}{!}{
    \begin{tabular}{l | ccccccccccc | cc}
    \toprule
         & \multicolumn{13}{c}{\textbf{mAP} (\%) at 10cm $\uparrow$} \\
        \midrule
        %\textbf{Joint} & RF~\cite{shotton2012efficient} & IEF~\cite{carreira2016human} & VI~\cite{haque2016towards} & RTW~\cite{yub2015random} & CMB~\cite{wang2018convolutional} & REN-9x6x6~\cite{guo2017towards} & A2J~\cite{xiong2019a2j} & V2V$^*$~\cite{moon2018v2v} & DECA-D3~\cite{garau2021deca} & WSM~\cite{zhang2020weakly} & AdaPose~\cite{zhang2021sequential} & \textit{w/o forecasting} & \textit{Ours} \\
         & RF & IEF & VI & RTW & CMB & REN-9x6x6 & A2J & V2V$^*$ & DECA-D3 & WSM & AdaPose & \textit{Ours} &  \\
\textbf{Joint} & \cite{shotton2012efficient} & \cite{carreira2016human} & \cite{haque2016towards} & \cite{yub2015random} & \cite{wang2018convolutional} & \cite{guo2017towards} & \cite{xiong2019a2j} & \cite{moon2018v2v} & \cite{garau2023capsules} & \cite{zhang2020weakly} & \cite{zhang2021sequential} & \textit{w/o forecasting} & \textit{Ours} \\
        \midrule
        Head & $63.8$ & $96.2$ & $98.1$ & $97.8$ & $97.7$ & $98.7$ & $98.5$ & $98.3$ & $93.9$ & $98.1$ & $98.4$ & $98.9$ & $98.6$ \\
        Neck & $86.4$ & $85.2$ & $97.5$ & $95.8$ & $98.5$ & $99.4$ & $99.2$ & $99.1$ & $97.9$ & $99.5$ & $98.7$ & $99.0$ & $99.4$ \\
        Shoulders & $83.3$ & $77.2$ & $96.5$ & $94.1$ & $75.9$ & $96.1$ & $96.2$ & $97.2$ & $95.2$ & $94.7$ & $95.4$ & $97.5$ & $97.6$ \\
        Elbows & $73.2$ & $45.4$ & $73.3$ & $77.9$ & $62.7$ & $74.7$ & $78.9$ & $80.4$ & $84.5$ & $82.8$ & $90.7$ & $84.4$ & $84.4$ \\
        Hands & $51.3$ & $30.9$ & $68.7$ & $70.5$ & $84.4$ & $55.2$ & $68.3$ & $67.3$ & $56.5$ & $69.1$ & $82.1$ & $76.8$ & $77.4$ \\
        Torso & $65.0$ & $84.7$ & $85.6$ & $93.8$ & $96.0$ & $98.7$ & $98.5$ & $98.7$ & $99.0$ & $99.7$ & $99.7$ & $98.7$ & $98.8$ \\
        Hips & $50.8$ & $83.5$ & $72.0$ & $80.3$ & $87.9$ & $91.8$ & $90.8$ & $93.2$ & $97.4$ & $95.7$ & $96.4$ & $87.6$ &  $90.4$ \\
        Knees & $65.7$ & $81.8$ & $69.0$ & $68.8$ & $84.4$ & $89.0$ & $90.7$ & $91.8$ & $94.6$ & $91.0$ & $94.4$ & $86.8$ &  $89.7$ \\
        Feet & $61.3$ & $80.9$ & $60.8$ & $68.4$ & $83.8$ & $81.1$ & $86.9$ & $87.6$ & $92.0$ & $89.9$ & $92.8$ & $75.3$ &  $88.0$ \\
        \midrule
        Upper body & $70.7$ & $61.0$ & $84.0$ & $84.8$ & $80.6$ & $-$ & $-$ & $-$ & $83.0$ & $-$ & $-$ & $90.3$ & $90.4$ \\
        Lower body & $59.3$ & $82.1$ & $67.3$ & $72.5$ & $86.5$ & $-$ & $-$ & $-$ & $95.3$ & $-$ & $-$ & $85.5$ & $90.7$ \\
        \midrule
        %Total body & $65.8$ & $71.0$ & $77.4$ & $80.5$ & $83.4$ & $84.9$ & $88.0$ & $88.7$ & $88.7$ & $89.6$ & $93.4$ & $84.2$ &  $86.6$ \\
        %Total body (vis.) & $-$ & $-$ & $-$ & $-$ & $-$ & $-$ & $-$ & $-$ & $-$ & $-$ & $-$ & $88.0$ & $90.6$ \\
        \textbf{Total body} & $65.8$ & $71.0$ & $77.4$ & $80.5$ & $83.4$ & $84.9$ & $88.0$ & $88.7$ & $88.7$ & $89.6$ & $\mathbf{93.4}$ & $88.0$ & $\underline{90.6}$ \\
        \bottomrule
    \end{tabular}
    }
    \end{center}
    \caption{Per-joint results on human pose estimation on ITOP side-view test set. The best result is reported in bold, while the second best is underlined. 
    As shown, the proposed framework achieves a significant accuracy on the total body, even though not expressively developed for the HPE task.
    Method marked with $^*$ uses 10 models ensemble.}
    \label{tab:itop-3d-results_joint}
\end{table*}

\begin{figure}[t]
    \centering
    \includegraphics[width=0.95\columnwidth]{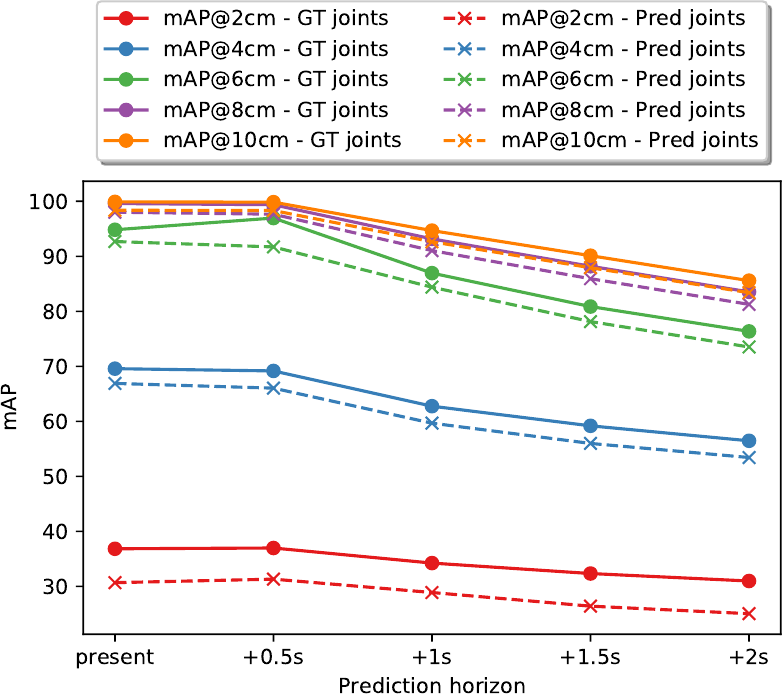}
    \caption{Comparison on Simba in terms of mAP using ground truth and predicted 3D joints as input to pose forecasting branch.}
    \label{fig:simba_forecasting_map}
\end{figure}

\subsection{Results}
We report results on SimBa and ITOP, both with our full pipeline and with a baseline not leveraging the nowcasting paradigm.
%either in single-frame mode, \ie only using a depth image as input and in multi-frame mode, \ie exploiting past predicted joint positions and then embedding the temporal information. 
In all experiments, when the model is optimized to forecast the future, past poses are fed at 10Hz for a duration of 1s. In output instead, we sample poses at 2Hz with a temporal horizon of 2s maximum.

% With SimBa, as competitors, we select four methods belonging to different scenarios usually exploited to predict the  pose in the 3D space.
% Firstly, we adopt a ResNet-18 architecture trained to directly regress 3D coordinates starting from depth maps as input. 
% As the second competitor, we evaluate the method proposed in~\cite{martinez2017simple}, consisting of a sequence of several MLPs trained to predict 3D joint coordinates relying on their 2D positions. We observe that this approach only provides relative joint locations with respect to a specific root (in our case the robot base). 
% The third competitors is based on the volumetric heatmaps

% \input{iccv2023AuthorKit/tables/itop_test}

%RV propongo di dividere i risultati su simba e itop, più che RPE e RPF...
%\subsubsection{RPE}
%\subsubsection{RPF}
\textbf{Results on SimBa.}
Table~\ref{tab:simba-3d-results} shows results on the SimBa dataset, reporting mean Average Precision (mAP) using different thresholds ($\delta = \{2, 4, 6, 8, 10\}$ cm) as well as ADD. 
%As mentioned above, the mean Average Precision (mAP) using different thresholds ($\delta = \{2, 4, 6, 8, 10\}$ cm) is reported in the central part of the table, while the ADD metric is shown on the last column.
%The performance provided by state-of-the-art approaches is given as reference. 
We report results using only the depth image (\textit{Ours w/o forecasting}) and with the additional input of past predicted 3D joints (\textit{Ours}).
% TODO: due parole sui competitors?
Following~\cite{simoni2022semi}, we test the same competitors to predict the 3D poses reporting the results in Table~\ref{tab:simba-3d-results}.
In particular, we train a ResNet-18~\cite{he2016deep} to directly regress 3D coordinates from depth maps.
We then evaluate the method proposed in~\cite{martinez2017simple}, a sequence of MLPs trained to estimate 3D joint coordinates relying on their 2D positions. This approach only provides relative joint locations with respect to a specific root (the robot base).
The third competitor, is based on the volumetric heatmap approach \cite{pavlakos2017coarse}, a representation for encoding 3D locations in a sampled 3D volume. This approach, in addition to a limited accuracy, leads to a significant video memory occupation of about 16GB, considerably higher than all the other methods (approximately 9 times higher than ours, see Section \ref{sec:time}).
Finally, \cite{simoni2022semi} uses the SPDH representation with a standard CNN.
Even without the use of the GRU input our approach yields the state of the art on SimBa. Interestingly, when exploiting past joints' locations with a recurrent network and adding the pose forecasting branch, results are improved further especially at low spatial thresholds, almost doubling mAP at the 2cm mark.

\begin{figure*}
    \centering
    \includegraphics[width=.88\textwidth]{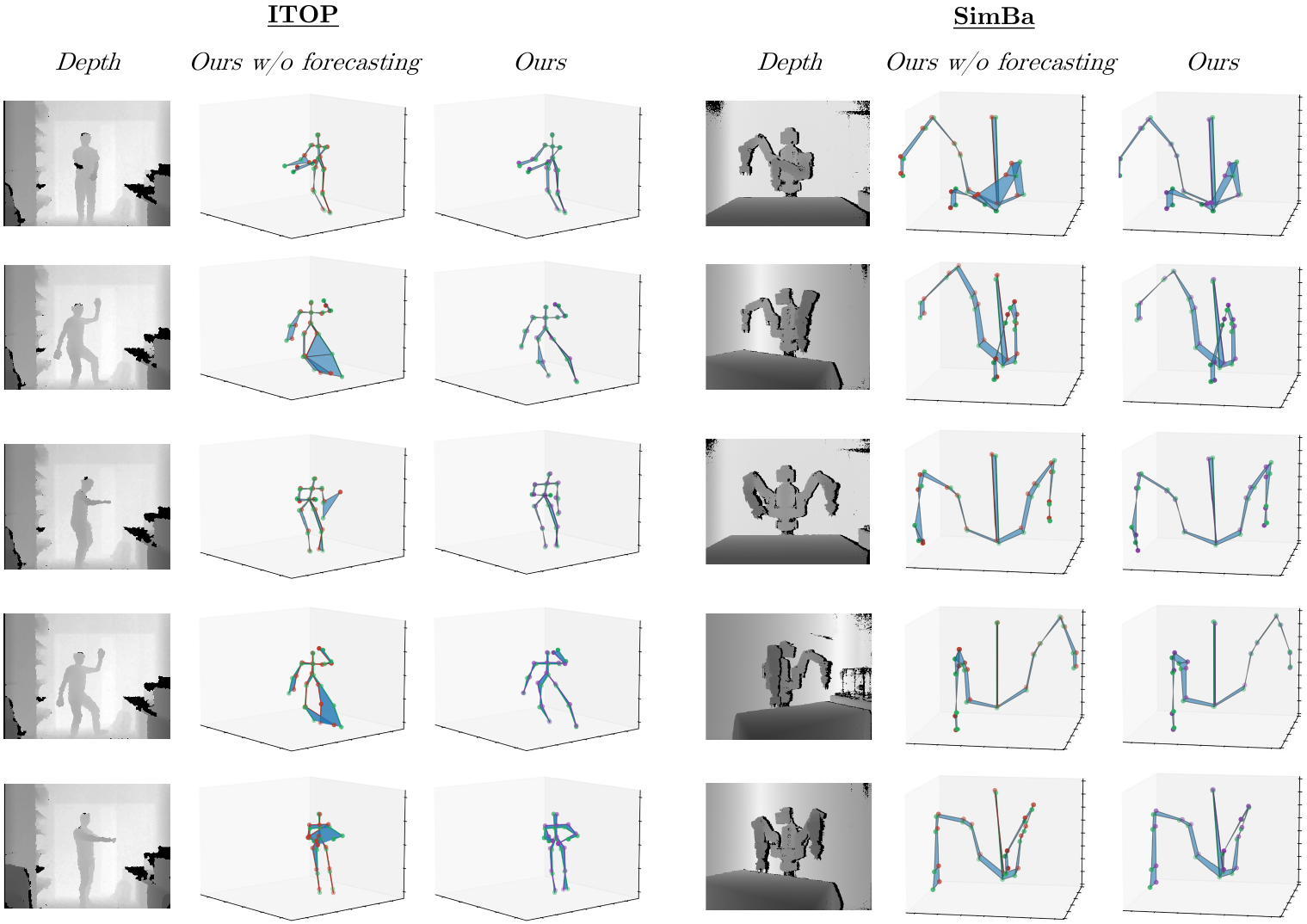}
    \caption{Qualitative examples for both ITOP and SimBa datasets where it can be appreciated the improvement in the pose estimation using the proposed approach. Green joints represent the ground truth pose, whereas red and violet represent respectively the poses estimated by our method without future and our full method. Blue regions connect ground truth skeletons and predictions, highlighting errors.}
    \label{fig:qualitative}
\end{figure*}

%\paragraph{Pose Forecasting}
Then, we show in Figure \ref{fig:simba_forecasting_map} the results for 3D Pose Forecasting by comparing mAP at different future timestamps. As an upper bound, we report results relying on ground truth past joints' locations. Interestingly, even when autoregressively feeding back estimated joints as input, the performance drop is limited with a maximum difference of $6\%$ for the 2cm threshold. Finally, as shown in Table~\ref{tab:simba-all-results}, it must be noted that at $1s$ ADD is roughly 1cm higher than the ADD at the current timestep prediction, making the approach suitable for collision detection.
Table~\ref{tab:simba-all-results} also shows a comparison between a simple baseline made of a linear regressor trained with SGD and our model with only the encoder-decoder for the forecasting branch. In the latter, the HRNet backbone extracting information from depth images is not used. In both configurations, we obtain much worse results, indicating the non-triviality of the task.
In Figure \ref{fig:qualitative} (right) we show qualitative results for poses predicted by our model with and without the forecasting branch, highlighting its importance.
\begin{table}[t]
    \begin{center}
    \resizebox{\columnwidth}{!}{
    \begin{tabular}{l ccccc c}
    \toprule
        & \multicolumn{5}{c}{\textbf{mAP} (\%) $\uparrow$} &  \\
        \cmidrule{3-7}
        \textbf{Horizon} & $\mathbf{2}$cm & $\mathbf{4}$cm & $\mathbf{6}$cm & $\mathbf{8}$cm & $\mathbf{10}$cm & \textbf{ADD} (cm) $\downarrow$ \\
        \midrule
        % Depth + 3D joints & \textit{Ours} & $t$ (vis. joints) & & $10.97$ & $41.53$ & $68.54$ & $83.62$ & $90.60$ & & $5.71$ \\
        $t$ & $10.19$ & $38.76$ & $64.32$ & $79.12$ & $86.57$ & $6.49$ \\
        % Depth + 3D joints & \textit{Ours} & $t+0.5s$ (all frames) & & $1.75$ & $8.44$ & $18.48$ & $29.21$ & $38.74$ & & $19.33$ \\
        % Depth + 3D joints & \textit{Ours} & $t+1s$ (all frames) & & $0.97$ & $5.72$ & $13.43$ & $22.67$ & $31.62$ & & $21.11$ \\
        $t+0.5s$ & $1.94$ & $9.61$ & $21.48$ & $33.91$ & $44.75$ & $17.66$ \\
        $t+1s$ & $1.20$ & $6.72$ & $16.39$ & $27.78$ & $38.56$ & $18.94$ \\
        % \midrule
        % All Image & SPDH (double branch + upsampling) & Curr (all joints) & & $8.99$ & $35.44$ & $59.38$ & $74.20$ & $82.43$ & & $7.86$ \\
        % All Image & SPDH (double branch + upsampling) & Curr (visible joints) & & $$ & $$ & $$  $$ & $86.39$ & & $6.98$ \\
        %  \midrule
        %  All Image + 3D joints & SPDH (concat+Current+Future) & Curr & & $6.69$ & $30.82$ & $55.47$ & $71.05$ & $80.60$ & & $8.18$ \\
        %  All Image + 3D joints & SPDH (concat+Current+Future) & +1s & & $0.39$ & $2.65$ & $6.89$ & $12.35$ & $18.65$ & & $23.62$ \\
        %  \midrule
        %  All Image + 3D joints & SPDH (concat+Current+Future) & Curr & & $$ & $$ & $$ & $$ & $$ & & $$ \\
        %  All Image + 3D joints & SPDH (concat+Current+Future) & +0.5s & & $$ & $$ & $$ & $$ & $$ & & $$ \\
        %  All Image + 3D joints & SPDH (concat+Current+Future) & +1s & & $$ & $$ & $$ & $$ & $$ & & $$ \\
        \bottomrule
    \end{tabular}
     }
    \end{center}
    \caption{Results on human pose estimation and forecasting on ITOP side-view test set. The model takes as input both depth and past poses. %The proposed method is tested in an autoregressive manner, taking as input the predicted 3D poses of the past $M$ timesteps in the more challenging human scenario.
    }
    \label{tab:3d-results}
\end{table}

\textbf{Results on ITOP.}
We show in Table~\ref{tab:itop-3d-results_joint} our results compared to the state-of-the-art.
Overall results for all methods on ITOP are generally worse than on SimBa, due to the fact human movements are more erratic and complex with respect to robot arm motion. Moreover, training is made more challenging by the presence of invalid joints, \ie joints without any manual annotation in the dataset. Nonetheless, on average considering the total body, our approach using a single depth frame is on par with most competing methods. Adding the supervision on future timesteps we rank above all methods except for AdaPose \cite{zhang2021sequential}, an approach expressively developed for the HPE task (differently from ours) which obtains a slightly higher mAP metric.

Furthermore, it is interesting to notice which joints benefit the most from nowcasting, \ie adding the forecasting branch. In general, the lower body registers a considerable improvement between the two variants of our approach. Hips and knees report a gain of approximately $+3\%$ mAP, whereas feet even $+13\%$ mAP.
Given that feet demonstrate greater dynamism in comparison to other body joints, they manifest behavior that is comparatively less erratic than, for instance, hands, wherein the advantageous outcome is less apparent.

In Table~\ref{tab:3d-results} we show the performance of the framework addressing the forecasting task, which is more challenging in the presence of wide movements performed by humans. These results can be a useful baseline reference for future works that address the forecasting task on ITOP.
In Figure~\ref{fig:qualitative} (left) we show qualitative results on ITOP, comparing the model with the present-only baseline.

\subsection{Execution Time Analysis}
\label{sec:time}
Our model must be deployable in a work environment, thus must be efficient for safety applications, \eg avoiding collisions and hazards.
We measured inference time on an Intel i7 ($2.90$ GHz) CPU and Nvidia Titan XP GPU.
The pose estimation branch alone runs at $20$ FPS. Adding the forecasting branch, observing autoregressively generated poses and estimating future ones, the overall inference time is around $11$ FPS with a video memory occupation of about $1.8$GB. Since we feed to the architecture 1 second of 3D poses sampled at $10$Hz and estimated by the model itself, we can run the whole framework in real-time without delays. The reaction time after observing the present frame before estimating the current and future poses is 90ms.

\section{Conclusion and Future Work}
We introduced the paradigm of 3D Pose Nowcasting, using depth data. The proposed framework jointly optimizes pose estimation and forecasting, exploiting two branches and the SPDH intermediate representation. We obtain state-of-the-art results in predicting current and near-future robot poses. The framework is also able to work with humans, achieving performance comparable with the current literature competitors on ITOP.
In future work, we plan to adopt Domain Adaptation techniques to reduce the Sim2Real shift, and the use of recent transformer-based architectures to model the input sequences.
Finally, we highlight the lack of depth-based datasets regarding human-machine interaction in social and working scenarios. This kind of data could lead to the realization of real-world collision detection and anticipation systems.

{\small
\bibliographystyle{ieee_fullname}
\bibliography{ref}
}

\end{document}